\documentclass[letterpaper, 10 pt, conference]{ieeeconf}
\IEEEoverridecommandlockouts
\usepackage{amsmath}
\usepackage{amssymb}
\usepackage{booktabs}
\usepackage{graphicx}
\usepackage{multirow}
\usepackage{url}
\usepackage{xcolor}
\usepackage{cite}

\begin{document}

\title{\LARGE \bf Runtime Safety Filtering for Two-Terminal Hazards\\in Robotic Battery Recycling}

\author{Yuxin Cao$^{1}$, Wei Song$^{2}$, Xianglin Yang$^{1}$, Fusen Guo$^{3}$, Lin Li$^{4}$, Xiao Cheng$^{5}$, Jin Song Dong$^{1}$\\[2pt]
{\normalsize $^{1}$National University of Singapore, Singapore \quad $^{2}$Griffith University, Australia \quad $^{3}$UNSW Canberra, Australia}\\
{\normalsize $^{4}$Southern Cross University, Australia \quad $^{5}$Macquarie University, Australia}
}

\maketitle

\begin{abstract}
Runtime safety filters for learned manipulation policies typically define unsafe states as unions of object-wise keep-out regions. This representation can be unnecessarily restrictive for hazards that depend on a joint spatial relation, such as battery recycling, where a conductive payload can short a charged cell only when it approaches both terminals simultaneously. We study runtime filtering for this two-terminal hazard in LIBERO using frozen OpenVLA policies. We factor a runtime filter into three design choices: the predicate structure, its geometric margin, and the fallback action applied when a commanded action is rejected. We compare a conjunctive predicate, a conventional two-site keep-out, and a composite of the two. For each predicate, we vary its margin to obtain a frontier between task success and residual hazard. We then compare four fallback strategies at matched operating points: holding, retreat, sampled search, and a continuous-action barrier projection. Across three workcells, the three predicate families trace nearly identical safety--utility frontiers once each is evaluated over its own margin. In contrast, the fallback strategy has a substantially larger effect: holding reduces task success by up to 0.302 relative to retreat without reducing hazard, while both minimally invasive fallbacks leave substantially more residual hazard. This ordering transfers to a second policy and task suite, while retreat-based filtering remains effective under standing errors in the clearances available to the filter, although correlated error in the estimated payload size is more damaging than larger independent errors in terminal position. These results show that, for proximity-defined manipulation hazards, margin selection and fallback strategy can matter more than predicate structure in determining the safety--utility trade-off of a runtime filter.
\end{abstract}

\begin{figure*}[t]
\centering
\includegraphics[width=0.98\textwidth]{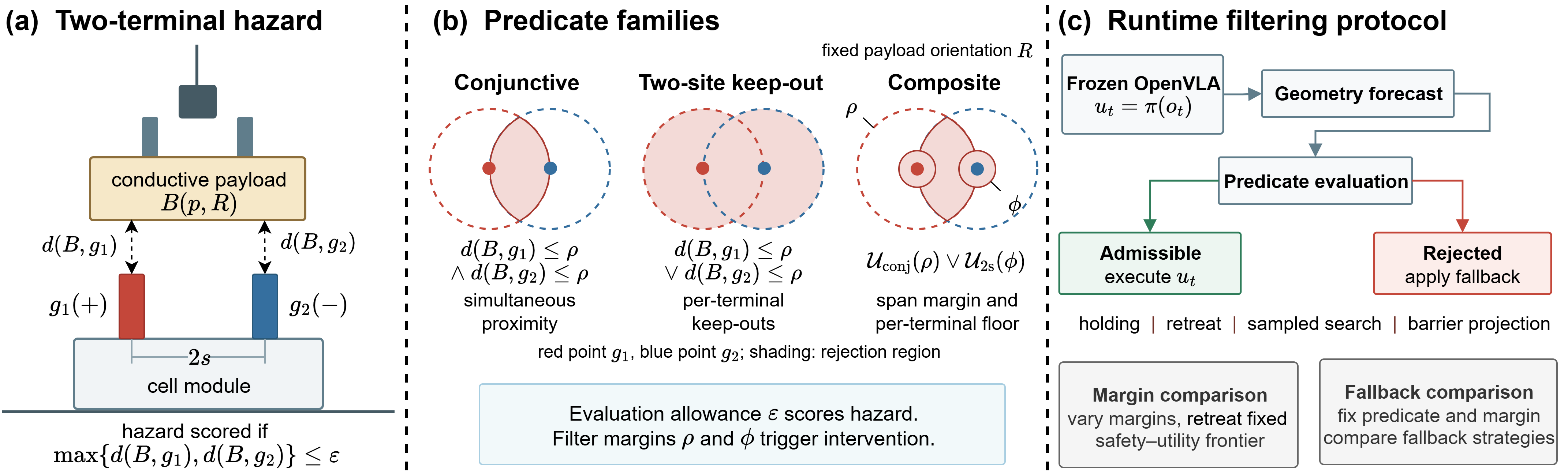}
\caption{Overview of two-terminal hazard evaluation and runtime filtering. (a) A conductive payload is scored hazardous when it lies within allowance $\varepsilon$ of both terminals. (b) Three predicate families define alternative rejection regions using margins $\rho$ and $\phi$. (c) The filter evaluates forecast geometry and either executes $u_t$ or applies a fallback.}
\label{fig:overview}
\end{figure*}

\section{Introduction}

Runtime safety filters provide a practical layer between learned manipulation policies and robot execution by rejecting or modifying actions that violate safety constraints~\cite{hsu2024safetyfilter}. In recent vision-language-action (VLA) systems, safety is commonly evaluated through unsafe instructions, collisions, contacts, object displacement, or interactions with hazardous objects~\cite{fan2026safevlabench,zhang2026guardianbench,chen2026hazardarena}. These hazards can often be represented using object-wise constraints, where the unsafe set is a union of keep-out regions around individual objects.

Battery energy storage is being deployed at scale, and the cells that reach end of life create a growing
recycling workload~\cite{mo2025energy,rana2022review}. Battery recycling introduces a different geometric hazard. A conductive payload can short-circuit a charged cell when it approaches both terminals simultaneously, whereas proximity to either terminal alone does not constitute the same bridging fault. Electrical safety standards similarly characterize bridging between conductive parts at different potentials through their relative clearances and separations~\cite{nfpa70e,iec60664}. The resulting unsafe set is therefore an intersection of two proximity conditions rather than a union of independent keep-outs. This distinction is relevant to robotic battery handling, where charged components and exposed conductive parts can coexist in the manipulation workspace~\cite{maddahi2015liveline,liu2025disassembly,hu2026second}.

Representing the physical hazard, however, is only one part of runtime filtering. A filter requires a margin that determines when intervention occurs and a fallback strategy that determines the action executed when the command is rejected. These choices are coupled: comparing two predicates at a shared margin can confound predicate structure with conservatism, while different fallback actions can change both task completion and residual hazard. We therefore study a runtime filter through three design choices: \emph{predicate structure}, \emph{margin}, and \emph{fallback strategy}. Our focus is how these choices jointly shape the empirical safety--utility trade-off of a learned manipulation policy.

We construct battery-recycling workcells in LIBERO and evaluate frozen OpenVLA policies under three predicate families: a conjunctive predicate that directly represents the two-terminal condition, a conventional two-site keep-out, and a composite of the two. We first vary the margin of each predicate and compare the resulting trade-off between task success and residual hazard. We then hold the predicate and margin fixed and compare four fallback strategies: holding, retreat, sampled search, and a continuous-action barrier projection. We additionally study what makes the hazard reachable, whether contact-based monitoring can detect it, and how behavior changes under a second policy and task suite, multiple modules, held-out layouts, and biased clearance estimates.

Across three workcells, the three predicate families trace nearly identical safety--utility frontiers once each is evaluated over its own margin. The fallback strategy has a substantially larger effect: holding reduces task success by 0.281--0.302 relative to retreat at the same predicate and margin without reducing hazard, while both minimally invasive fallbacks leave more residual hazard. We further find that hazard reachability depends strongly on terminal height and whether the policy transports the payload across the terminal pair, and that contact-based monitoring misses most events defined at positive clearance. The ordering of fallback strategies transfers to a second policy and task suite, and the retreat-based filters remain effective under standing clearance errors.

The contributions of this paper are summarized as follows:
\begin{itemize}
    \item We formulate two-terminal bridging as a proximity-defined manipulation hazard and study filtering through predicate structure, margin, and fallback strategy.
    
    \item We show that the predicate families trace similar safety--utility frontiers when each is evaluated over its own range of margins, while the fallback strategy has a substantially larger effect on safety and task success.
    
    \item We characterize hazard reachability and evaluate the observed trends across a second policy, additional workcells, layouts, and sensing errors.
\end{itemize}

\section{Related Work}

\noindent\textbf{Runtime safety filtering.}
Runtime safety filters for learned controllers have been studied through several formal frameworks. Predictive safety filters determine whether a proposed input admits a feasible trajectory back to a safe set over a finite horizon~\cite{wabersich2021predictive}. Backup control barrier functions construct invariant safe sets using an explicit backup controller~\cite{chen2021backup}, while Hamilton--Jacobi reachability characterizes states from which safety can be maintained against worst-case evolution~\cite{bansal2017hj}. These approaches and their relationships are reviewed in~\cite{hsu2024safetyfilter}. 
Our study focuses on the geometry-based end of this design space and examines how predicate structure, margin, and fallback strategy affect the safety--utility trade-off of a frozen learned manipulation policy.

\noindent\textbf{Safety for vision-language-action policies.}
Safety research for VLA models has largely focused on the policy itself. Recent benchmarks evaluate unsafe instructions, hazardous object interactions, and contextual risks in embodied tasks~\cite{fan2026safevlabench,zhang2026guardianbench,chen2026hazardarena}. Safety-aware finetuning incorporates safety constraints during policy learning~\cite{zhang2025safevla}, while runtime approaches use internal attention or action representations to identify and suppress unsafe actions~\cite{park2026attentionfilter,beaudin2026anybodyguard}. The broader threat and mitigation landscape is summarized in~\cite{vlasafety2026survey}. In contrast, we keep the VLA policy frozen and study an external runtime filter whose intervention is determined by workcell geometry. The hazard also differs from those commonly evaluated in current VLA benchmarks: it can occur under a nominal task instruction and at positive clearance, without requiring contact, collision, or an explicitly unsafe command.

\noindent\textbf{Electrical hazards and robotic manipulation.}
A conductive object bridging two points at different electrical potentials is a well-established electrical safety hazard. Live-line practice and insulation standards specify clearance and creepage requirements for energized equipment~\cite{nfpa70e,iec60664}. Related work has considered robotic manipulation around energized infrastructure, the disassembly of electrical components, and the identification and health assessment of retired cells~\cite{maddahi2015liveline,liu2025disassembly,hu2026second,guo2025battery}. These works motivate geometric safeguards around hazardous components and characterize the condition of the cells being handled, but they do not study how a learned manipulation policy interacts with a two-terminal bridging condition. Our setting connects this physical hazard to runtime filtering for VLA-controlled manipulation.

\noindent\textbf{Compositional safety constraints.}
The ability to represent conjunctions and other logical combinations of safety constraints is well established. Nonsmooth control barrier functions compose constraints through minimum and maximum operators~\cite{glotfelter2017nonsmooth,glotfelter2018boolean,glotfelter2019hybrid}, while temporal-logic formulations provide related mechanisms for composing specifications~\cite{lindemann2019stlcbf}. Runtime shielding has also been used to enforce safety constraints around learned policies~\cite{alshiekh2018shielding}, and semantic and relational safeguards have been studied for robot manipulation~\cite{brunke2025semantic}. Building on these established compositional formulations, we study the empirical consequence of predicate choice: whether a predicate that directly matches the two-terminal hazard yields a better safety--utility trade-off when margin and fallback strategy are controlled.

\section{Runtime Filter Design}
\label{sec:system}

\subsection{Problem Formulation}
\label{sec:hazard}
At control step $t$, let a frozen manipulation policy $\pi$ map the current observation $o_t$ to a proposed action $u_t=\pi(o_t)$. A runtime filter then evaluates this proposed action using a safety predicate and either executes the original command or replaces it with a suitable fallback action. We therefore characterize the runtime filter by three design choices: \emph{predicate structure}, \emph{margin}, and \emph{fallback strategy}.

We denote the carried body at position $p$ and orientation $R$ as $B(p,R)$, and the two terminals separated by $2s$ along their connecting axis as $g_1$ and $g_2$, as shown in Fig.~\ref{fig:overview}(a). The clearance between the carried body and a terminal is denoted as $d(B,g)$. At margin $\rho$, a conventional two-site keep-out excludes
\begin{equation}
\mathcal{U}_{\mathrm{2s}}(\rho)=\{(p,R):d(B,g_1)\le\rho\ \vee\ d(B,g_2)\le\rho\},
\label{eq:two_site}
\end{equation}
whereas a predicate that directly captures the two-terminal condition excludes
\begin{equation}
\mathcal{U}_{\mathrm{conj}}(\rho)=\{(p,R):d(B,g_1)\le\rho\ \wedge\ d(B,g_2)\le\rho\}.
\label{eq:conj}
\end{equation}
Thus, $\mathcal{U}_{\mathrm{2s}}(\rho)\setminus\mathcal{U}_{\mathrm{conj}}(\rho)$ is restricted by the conventional keep-out although it does not satisfy the two-terminal condition. The evaluation later uses a separate hazard allowance, so the filter margin and hazard criterion are not conflated.

The difference depends on both terminal spacing and payload geometry. Along the terminal axis, let $w$ denote the payload half-extent. The conjunctive region is reachable only when $w+\rho\ge s$, and can therefore disappear for a narrow payload at a small margin while the two-site keep-out remains nonempty. Conversely, once $\rho\ge s$, the conjunctive condition no longer requires finite payload extent. Fig.~\ref{fig:overview}(b) illustrates the two regions and their difference.

Payload orientation changes its effective span along the terminal axis. A fixed workspace keep-out therefore cannot generally reproduce $\mathcal{U}_{\mathrm{conj}}$ for a rotating payload. This dependence motivates evaluating each predicate over its own margin values instead of comparing rules at a single shared margin. The overall runtime filtering protocol is summarized in Fig.~\ref{fig:overview}(c), from geometry forecasting and predicate evaluation to command execution or fallback selection.

\subsection{Composite Predicate}
\label{sec:composite}

The conjunctive predicate captures simultaneous proximity to both terminals but imposes no minimum clearance from either terminal individually. This may be undesirable under disturbances or with several modules. The two-site keep-out has the opposite limitation: it maintains clearance from every terminal but can unnecessarily restrict task-relevant motion.

We therefore combine the two constraints. Let $\hat{p}_{t+k}$ denote the forecast payload position under the commanded action and $d_j$ the clearance to terminal $j$. Over a horizon of $H$ control steps, the composite predicate admits an action when
\begin{equation}
\min_{k\le H}\max_j d_j(\hat{p}_{t+k})>\rho \quad \wedge \quad \min_{k\le H}\min_j d_j(\hat{p}_{t+k})>\phi,
\label{eq:composite}
\end{equation}
where $\rho$ is the margin for simultaneous proximity and $\phi$ is a smaller per-terminal clearance. The first term limits approaches to both terminals, while the second maintains clearance from each terminal individually. In multi-module workcells, the first term is evaluated for each module's physical terminal pair, while the per-terminal floor is applied to all terminals. An action is admissible only when all resulting constraints are satisfied. The commanded translation is propagated over this horizon with orientation held fixed, and all predicate variants use the same forecasted geometry.

Disabling the second term recovers the conjunctive predicate, while setting $\phi=\rho$ recovers the two-site keep-out. The composite therefore connects the two predicate structures within a common family and enables intermediate operating points to be compared directly without changing the underlying forecast model or geometry.

\subsection{Fallback Strategies}
\label{sec:lr}

When the commanded action violates the active predicate, the filter must select a fallback action. We compare four strategies with the predicate and margin held fixed. \emph{Holding} removes the commanded translation. \emph{Retreat} selects an admissible direction that maximizes predicate slack and moves the payload away from the active constraint. \emph{Sampled search} selects the first admissible candidate from a fixed set ordered by deviation from the policy command, ranging from scaled commands through blends with an escape direction to the escape direction itself.

The fourth strategy, \emph{barrier projection}, searches the continuous action space for an admissible correction that remains close to the commanded action. Let $u$ denote the commanded translation, $h$ the predicate slack, and $h_0$ the current-state slack before applying the correction. It solves
\begin{equation}
\min_{u'}\;\lVert u'-u\rVert^2 \quad \text{subject to} \quad h(u')\ge(1-\alpha)h_0,
\label{eq:projection}
\end{equation}
where $\alpha$ controls how much of the available slack the corrected action may consume. We locally linearize the constraint, project along the correction direction, and use bisection to move the correction toward the original command while preserving admissibility. If no correction satisfies the required improvement, the filter falls back to holding.

Because the conjunctive predicate contains a maximum over terminal clearances, this procedure computes a local correction without guaranteeing global optimality. We therefore use it as a projection baseline based on the barrier constraint and interpret the resulting correction as approximate.

\subsection{Baselines}
\label{sec:baselines}

The main comparison includes the conjunctive predicate, the two-site keep-out, and the composite predicate. We consider three geometric baselines: a keep-out around one terminal, a keep-out centered at the midpoint of the pair, and a box spanning the inter-terminal gap.

We treat the margin as a predicate-specific parameter that can vary across rules. The same numerical margin can impose different geometric restrictions under different predicate structures and workcell configurations. We therefore evaluate each predicate over its own range of margins and compare the resulting safety--utility trade-offs. For fallback comparisons, the predicate and margin are fixed so that only the post-intervention action changes.

\section{Experiments}

\subsection{Experimental Setup}

\noindent\textbf{Policy and simulator.}
We evaluate OpenVLA-7B~\cite{kim2024openvla} finetuned on LIBERO-Object~\cite{liu2023libero} and keep the policy frozen throughout evaluation. The simulator is robosuite~\cite{zhu2020robosuite} with MuJoCo~\cite{todorov2012mujoco}, and the robot uses an operational space controller. The battery module replaces an existing distractor in the LIBERO scene instead of adding another object, preserving object count and clutter level seen during finetuning.

\noindent\textbf{Workcell geometry.}
The battery module is a rectangular case with two exposed busbars of opposite polarity. We vary both their separation and height above the bench. The primary carried conductor is a grocery item from LIBERO whose half-extent along the terminal axis ranges from 18.0~mm edge-on to 34.1~mm across its diagonal. For the standard 72~mm terminal pair, the 36~mm half-separation is therefore 1.9~mm beyond the maximum span of the carton, whereas the narrower pairs can be spanned at suitable orientations.

\noindent\textbf{Evaluation conditions.}
We organize the evaluation around six workcell conditions. The fixed benches place the module at a fixed pose and are used to measure excess-region occupancy and geometric baselines. The frontier bench uses a 50~mm terminal pair elevated to the payload transport height and positioned between the pick and place locations. This is the main condition for comparing margins and fallback strategies. The narrow-pair condition reduces the separation to 38~mm. The multi-module condition contains three modules and six busbars. The held-out condition samples a new module pose in each episode using margins selected elsewhere. Finally, the second-policy condition uses a LIBERO-Spatial checkpoint carrying a bowl on its native task.

\noindent\textbf{Metrics and statistics.}
We score a geometric hazard event at allowance $\varepsilon$ when the larger of the two terminal clearances reaches $\varepsilon$ or below during an episode. This allowance $\varepsilon$ is distinct from the filter margin $\rho$. Task success is the fraction of episodes that complete the assigned task, hazard rate is the fraction containing at least one geometric hazard event, and intervention rate is the fraction of control steps where the predicate rejects the policy action. These metrics are denoted as Success, Hazard, and Interv.\ in the tables. Compared arms use matched episode seeds. Episode outcomes are evaluated with McNemar's exact test~\cite{mcnemar1947} on discordant pairs. Unless stated otherwise, each comparison uses six runs of 32 episodes, giving 192 matched episodes.

\noindent\textbf{Filter implementation.}
All predicates use the same simulator geometry queries and a prediction horizon of $H=3$, with payload orientation fixed during the forecast. Unless otherwise stated, filtered experiments use retreat as the fallback. We verified the disabled-floor and $\phi=\rho$ endpoints of the composite predicate against independent implementations on 20,000 random states and observed no disagreement. For barrier projection, no direction satisfies the required slack improvement on 9.0\% of interventions. These cases revert to holding as defined in Section~\ref{sec:lr}.

\subsection{Hazard Reachability}
\label{sec:reach}
\noindent\textbf{Terminal height and transport path.}
The two-terminal hazard is not reachable in the default workcell. At bench height, the policy transports the carton above the terminals, and the minimum simultaneous clearance is 35.1~mm. Reducing the pair separation from 72~mm to 38~mm lowers this value to 20.1~mm but still produces no contact because the payload remains above the busbars as it crosses the pair.

Raising the terminals into the transport corridor changes this behavior. As shown in Table~\ref{tab:reach}, increasing the height for the carried can reduces the tightest simultaneous clearance from 50.7~mm at 119~mm height to penetration at 196~mm, where four of 192 episodes contact both busbars. Geometry alone is nevertheless insufficient to determine reachability. A rigid bar welded below the wrist has a support range of 6.0--55.3~mm and satisfies the span condition in Section~\ref{sec:hazard}, yet never approaches both terminals within 20~mm because the policy transports it with the wrist level. Similarly, the second policy remains outside the 250~mm measurement window when the module is placed beside its route but reaches within 20~mm in 0.094 of episodes when the same module is moved onto the transport path. These results identify transport path and terminal height as the main factors governing whether the geometric hazard is encountered.

\begin{table}[t]
\caption{Hazard reachability under the unfiltered policy.}
\label{tab:reach}
\centering
\small
\setlength{\tabcolsep}{4pt}
\begin{tabular}{@{}llrrr@{}}
\toprule
Conductor & Pair & Height & Success & Tightest Clearance \\
\midrule
carton & 72~mm & 45~mm & 0.875 & 35.1~mm \\
carton & 38~mm & 45~mm & 0.865 & 20.1~mm \\
\midrule
can & 72~mm & 119~mm & 0.714 & 50.7~mm \\
\textbf{can} & \textbf{72~mm} & \textbf{196~mm} & \textbf{0.422} & \textbf{$-$79.8~mm} \\
\midrule
welded bar & 50~mm & 196~mm & 0.552 & 21.8~mm \\
bowl & 50~mm & 196~mm & 0.867 & $>$250~mm \\
\bottomrule
\end{tabular}
\end{table}

\subsection{Predicate--Margin Trade-off}

\noindent\textbf{Excess-region occupancy.}
\label{sec:excess}
A more restrictive predicate matters only when nominal policy trajectories enter the additional region it excludes. We therefore run the policy unfiltered on the six fixed benches and evaluate all predicates offline at every control step. At a 20~mm margin, the conjunctive predicate activates on 0.2\% of steps, compared with 7.2\% for a single-terminal keep-out, 10.9\% for the two-site keep-out, and 2.5\% for the midpoint keep-out. At 40~mm, the corresponding rates are 3.7\%, 24.6\%, 41.9\%, and 23.1\%. Relative to the conjunctive predicate, the single-terminal and two-site rules therefore exclude an additional 7.0 and 10.7 percentage points of policy actions at 20~mm, increasing to 20.9 and 38.2 points at 40~mm. Under holding, these additional interventions directly reduce task completion.

\begin{figure*}[t]
\centering
\includegraphics[width=0.73\textwidth]{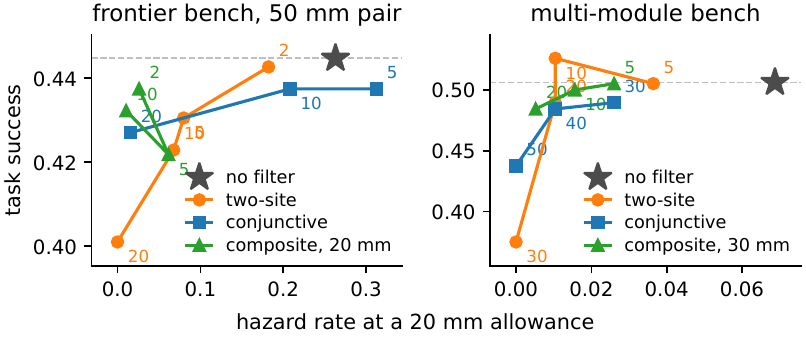}
\caption{Safety--utility frontiers obtained by varying the margin of each predicate. Labels denote margins in millimeters.}
\label{fig:curves}
\end{figure*}

\noindent\textbf{Geometric baselines.}
We next compare the conjunctive predicate with single-region alternatives on the fixed benches. Each filter uses retreat and is evaluated at a margin giving a comparable hazard rate. A keep-out spanning the gap between the terminals is the strongest of the four single-region baselines, retaining 0.370 task success compared with 0.604 for the conjunctive predicate. The midpoint keep-out is most restrictive, intervening on 41.6\% of policy actions to reach the same hazard level. Thus, a better geometric approximation to the bridging region reduces unnecessary intervention, although the comparison still depends on the selected margin.

\noindent\textbf{Evaluation allowance.}
\label{sec:degenerate}
The hazard allowance also affects interpretation of the metric. For a terminal pair with half-separation $s=36$~mm, a hazard event at allowance $\varepsilon$ requires a payload half-extent of at least $s-\varepsilon$. This is 16~mm at $\varepsilon=20$~mm and zero at $\varepsilon=40$~mm. At or above the half-separation, the metric measures simultaneous proximity independent of a finite spanning extent. Re-scoring the same episodes at 50~mm reverses the apparent ordering, with hazard rates of 0.547 for the conjunctive predicate and 0.438 for the two-site keep-out. We use the 20~mm allowance for the main frontier comparison, where the geometric differences between predicates remain meaningful.

\noindent\textbf{Safety--utility frontier.}
\label{sec:frontier}
A predicate with a free margin defines a family of operating points across filter settings. We therefore vary the margin of each predicate independently on the frontier bench with retreat fixed, and plot task success against residual hazard in the left panel of Fig.~\ref{fig:curves}. 
The three predicate families produce similar frontiers. The two-site keep-out at a 10~mm margin reaches a hazard rate of 0.016 with task success 0.443. The conjunctive predicate at 20~mm reaches the same hazard rate with success 0.427, while the composite with a 20~mm span margin and 10~mm floor reaches 0.010 with success 0.432. All three reduce the unfiltered hazard rate of 0.269 at $p<10^{-8}$. Their hazard outcomes differ by at most one or two episodes in 192, and their task-success reductions relative to the unfiltered policy are not statistically significant ($p=0.200$, $0.088$, and $0.099$).

The composite family contains the conjunctive and two-site predicates as endpoints, but its intermediate settings do not improve the frontier. A fixed-margin comparison suggests a different conclusion: at 5~mm, the conjunctive and two-site predicates yield hazard rates of 0.312 and 0.080, respectively. Once each predicate is evaluated at a margin suited to its geometry, both reach 0.016. The apparent advantage at a fixed margin is primarily a margin effect.

\noindent\textbf{Narrower terminal pair.}
\label{sec:micro}
Reducing the terminal separation from 50~mm to 38~mm leaves the hazard rate similar, 0.257 versus 0.269 on the frontier bench, and preserves the fallback ordering. With the composite, holding, retreat, and sampled search achieve task success of 0.125, 0.432, and 0.443 and hazard rates of 0.036, 0.016, and 0.130, respectively. The margin does not transfer across geometries. For the 38~mm pair, a 20~mm span margin covers the inter-terminal region. A coarse margin grid initially suggested a 0.024 task-success advantage for the composite because the two-site baseline was tested only at 5 and 10~mm, where its hazard rate changed from 0.120 to 0.021. Adding intermediate margins removes this apparent advantage, leaving a task-success difference of 0.016 at $p=0.581$.

\subsection{Fallback Strategies}
\label{sec:response}

\noindent\textbf{Main comparison.}
Table~\ref{tab:response} compares the four fallback strategies at matched predicate operating points: 20~mm for the conjunctive predicate, 10~mm for the two-site keep-out, and a 20~mm span margin with a 5~mm floor for the composite. Thus, only the fallback changes within each predicate. Holding produces the largest loss in task performance. Relative to retreat on matched episodes, task success decreases by 0.281 for the conjunctive predicate, 0.297 for the two-site keep-out, and 0.302 for the composite, with $p<10^{-14}$ in all three comparisons. These reductions provide no measurable improvement in hazard rate.

The two minimally invasive strategies show the opposite pattern. Relative to retreat, barrier projection increases the 20~mm hazard rate by 0.125, 0.141, and 0.089 across the three predicates, while sampled search increases it by 0.203, 0.328, and 0.148. All six hazard differences are statistically significant at $p<10^{-3}$. Neither strategy significantly improves task success over retreat. Relative to the unfiltered hazard rate of 0.269, retreat reduces hazard by approximately 0.20 for all three predicates, with $p$ between $5.5\times10^{-10}$ and $5.4\times10^{-9}$. Under barrier projection, the corresponding reductions are 0.063, 0.057, and 0.098, and the first two are not statistically significant ($p=0.071$ and $p=0.110$).

\begin{table}[t]
\caption{Fallback strategies on the frontier bench.}
\label{tab:response}
\centering
\small
\setlength{\tabcolsep}{4pt}
\begin{tabular}{@{}llrrr@{}}
\toprule
Predicate & Fallback & Success & Hazard & Interv. \\
\midrule
\multicolumn{2}{@{}l}{No filter} & 0.448 & 0.269 & --- \\
\midrule
\multirow{4}{*}{Conjunctive rule} & holding & 0.146 & 0.036 & 20.6\% \\
 & retreat & 0.427 & 0.016 & 2.0\% \\
 & barrier projection & 0.469 & 0.141 & 1.7\% \\
 & sampled search & 0.500 & 0.219 & 3.5\% \\
\midrule
\multirow{4}{*}{Two-site keep-out} & holding & 0.146 & 0.047 & 21.2\% \\
 & retreat & 0.443 & 0.016 & 1.7\% \\
 & barrier projection & 0.453 & 0.156 & 1.6\% \\
 & sampled search & 0.516 & 0.359 & 4.0\% \\
\midrule
\multirow{4}{*}{Composite} & holding & 0.137 & 0.039 & 22.0\% \\
 & retreat & 0.438 & 0.021 & 1.8\% \\
 & barrier projection & 0.438 & 0.109 & 1.9\% \\
 & sampled search & 0.391 & 0.164 & 3.4\% \\
\bottomrule
\end{tabular}
\end{table}

\noindent\textbf{Boundary behavior.}
The result is consistent with how the fallback strategies modify the policy command. A minimally invasive correction preserves as much of the commanded motion as possible and can keep the payload close to the active constraint boundary. For a hazard defined by proximity, repeated motion near that boundary increases residual hazard. The effect is not specific to discrete search or the barrier formulation. Modifying sampled search to reach the boundary directly rather than closing half of the gap changes the conjunctive hazard rate from 0.324 to 0.297, while relaxing the barrier projection from $\alpha=0.5$ to $\alpha=0.9$ changes the composite hazard rate from 0.109 to 0.177. These results support the boundary-proximity explanation.

\begin{table}[t]
\caption{Transfer to a second policy (LIBERO-Spatial) and task suite.}
\label{tab:second}
\centering
\small
\setlength{\tabcolsep}{4pt}
\begin{tabular}{@{}lrrr@{}}
\toprule
Filter & Success & Hazard & Interv. \\
\midrule
No filter & 0.594 & 0.094 & --- \\
Conjunctive, 20~mm, retreat & 0.625 & 0.031 & 0.3\% \\
Two-site, 10~mm, retreat & 0.562 & 0.062 & 1.3\% \\
Composite, 20 and 5~mm, retreat & 0.562 & 0.031 & 1.2\% \\
Two-site, 10~mm, hold & 0.562 & 0.094 & 3.7\% \\
Two-site, 10~mm, sampled search & 0.594 & 0.094 & 0.8\% \\
Two-site, 10~mm, barrier projection & 0.594 & 0.094 & 0.7\% \\
\bottomrule
\end{tabular}
\end{table}

\subsection{Generalization and Robustness}

\noindent\textbf{Carried objects.}
We repeat the predicate comparison with three additional carried objects at the 50~mm terminal pair, using 192 matched episodes per arm and a 10~mm evaluation allowance. The two-site keep-out reduces hazard from 0.182 to 0.000, from 0.203 to 0.005, and from 0.302 to 0.005 across the three objects, with $p$ between $5.8\times10^{-11}$ and $2.1\times10^{-16}$. It intervenes on only 2.3--2.5\% of actions and improves task success on two of the three objects. At the same 5~mm margin, the conjunctive predicate achieves roughly half of the hazard reduction, but this fixed-margin difference disappears once the predicates are evaluated over their own margin ranges, consistent with the main frontier result.

\noindent\textbf{Second policy.}
\label{sec:second}
We next evaluate a LIBERO-Spatial policy carrying a bowl on its native task, with the module placed on the route between the pick and place locations. The unfiltered policy achieves 0.594 task success and a 20~mm hazard rate of 0.094. Table~\ref{tab:second} reports 192 matched episodes per condition under the same evaluation protocol.

The fallback ordering transfers consistently to the second policy. Retreat reduces the hazard rate to 0.031 for both the conjunctive and composite predicates at $p=4.9\times10^{-4}$ and to 0.062 for the two-site keep-out at $p=0.031$. Holding leaves the two-site hazard rate unchanged at 0.094 while reducing task success by 0.031, and both minimally invasive strategies exactly match the unfiltered policy in both task success and hazard on the matched episodes. 
The predicate operating points do not align as closely as in the main experiment because their margins are transferred from the grocery-policy setting without retuning. In particular, the 10~mm two-site margin is the smallest of the three and provides less clearance for the new payload. This result further indicates that filter margins should be retuned when payload geometry changes.

\begin{table}[t]
\caption{Sensitivity to standing errors in the measured clearance.}
\label{tab:noise}
\centering
\small
\setlength{\tabcolsep}{4pt}
\begin{tabular}{@{}lrrr@{}}
\toprule
Filter & Success & Hazard & Interv. \\
\midrule
No filter & 0.455 & 0.269 & --- \\
Two-site, exact geometry & 0.443 & 0.016 & 1.7\% \\
Two-site, terminals 5~mm & 0.458 & 0.036 & 1.7\% \\
Two-site, terminals 10~mm & 0.427 & 0.073 & 1.8\% \\
Two-site, terminals 20~mm & 0.375 & 0.094 & 2.7\% \\
Two-site, payload 10~mm & 0.411 & 0.120 & 1.5\% \\
Composite, exact geometry & 0.438 & 0.021 & 1.8\% \\
Composite, terminals 5~mm & 0.432 & 0.036 & 1.7\% \\
Composite, terminals 10~mm & 0.432 & 0.078 & 1.9\% \\
Composite, terminals 20~mm & 0.396 & 0.083 & 2.4\% \\
Composite, payload 10~mm & 0.411 & 0.089 & 1.6\% \\
\bottomrule
\end{tabular}
\end{table}

\noindent\textbf{Sensing error.}
\label{sec:noise}
The preceding experiments provide the filter with exact simulator geometry. To test sensitivity to imperfect geometry, we perturb the clearances available to the filter while continuing to score hazard on the true geometry. A bias is sampled once per episode from a zero-mean Gaussian with standard deviation $\delta$ and held fixed throughout the episode, thereby modeling persistent calibration error. Terminal error uses independent biases for the two sites, whereas payload error applies one common bias to all terminal clearances. Table~\ref{tab:noise} reports both perturbations with retreat as the fallback and a 20~mm evaluation allowance.

A 5~mm terminal error increases the hazard rate from 0.016 to 0.036 for the two-site keep-out and from 0.021 to 0.036 for the composite, but neither change is statistically significant relative to exact geometry. At 20~mm, the rates rise to 0.094 and 0.083. For the two-site predicate, this error is twice its 10~mm margin, yet the filter reduces the unfiltered hazard rate from 0.269 to 0.094 at $p=0.001$, with task success decreasing from 0.443 to 0.375 as intervention rises from 1.7\% to 2.7\%.

Payload-size error is more damaging. A 10~mm common payload bias raises the two-site hazard rate to 0.120, exceeding 0.094 under a 20~mm terminal-position error, with fewer interventions. Independent terminal errors can make clearances appear smaller or larger, whereas an underestimated payload shifts all clearances in the same unsafe direction. The carried object's geometry is at least as important to the filter as localization of the terminal sites.

\noindent\textbf{Held-out layout.}
\label{sec:tight}
The main operating points are selected on workcells with fixed module placement. We therefore evaluate a held-out condition in which a new module pose is sampled for every episode and all margins are fixed from other conditions. The composite uses a 30~mm span margin and a 10~mm floor. 
Table~\ref{tab:tight} shows that all three predicates transfer to the new layouts. The conjunctive predicate reduces the 20~mm hazard rate from 0.115 to 0.021 at $p=7.6\times10^{-6}$, while the two-site and composite predicates reach 0.031 and 0.005, respectively. The filtered conditions differ from one another by only one to six episodes in 192. The fallback effect remains larger: holding with the composite reduces task success by 0.214 relative to the unfiltered policy at $p=1.0\times10^{-11}$, whereas retreat reduces it by 0.052. Sampled search recovers 0.041--0.052 task success relative to retreat across the predicate comparisons but consistently increases residual hazard.

\begin{table}[t]
\caption{Held-out random-layout evaluation.}
\label{tab:tight}
\centering
\small
\setlength{\tabcolsep}{4pt}
\begin{tabular}{@{}lrrr@{}}
\toprule
Filter & Success & Hazard & Interv. \\
\midrule
No filter & 0.755 & 0.115 & --- \\
Conjunctive, 30~mm, retreat & 0.740 & 0.021 & 1.0\% \\
Two-site, 10~mm, retreat & 0.714 & 0.031 & 1.0\% \\
Composite, hold & 0.542 & 0.021 & 16.5\% \\
Composite, retreat & 0.703 & 0.005 & 1.5\% \\
Composite, sampled search & 0.745 & 0.026 & 2.4\% \\
\bottomrule
\end{tabular}
\end{table}

\noindent\textbf{Multiple modules.}
\label{sec:dense}
We finally place three battery modules and six busbars in the same workcell. This setting is particularly challenging for the conjunctive predicate because maintaining adequate separation from one terminal pair can still place the payload near terminals belonging to another module. It also directly tests the per-terminal floor in the composite predicate. 
The right panel of Fig.~\ref{fig:curves} shows the margin-dependent frontiers. The three predicate families again interleave. The best individual operating point is the two-site keep-out at a 10~mm margin, which reduces hazard from 0.073 to 0.010 at $p=4.9\times10^{-4}$ while achieving task success 0.526 compared with 0.510 without filtering. At matched task success of 0.484, the conjunctive and composite predicates obtain hazard rates of 0.010 and 0.005, a difference of one episode in 192. All three families can reach zero hazard at more restrictive margins. Increasing the composite floor moves its frontier from the conjunctive endpoint toward the two-site endpoint but does not produce a new operating point that dominates both.

\subsection{Analysis and Cost}

\noindent\textbf{Margin--predicate interaction.}
\label{sec:diagnosis}
The conjunctive and two-site predicates behave differently at small margins but converge as their margins increase. At 5~mm, the conjunctive predicate allows the payload to travel farther across the terminal pair before intervention and produces a hazard rate of 0.312, whereas the two-site keep-out intervenes earlier and reaches 0.080. On these workcells, however, a trajectory that approaches both terminals first approaches at least one terminal. Once the two-site margin is large enough, it therefore intercepts the same hazardous episodes as the conjunctive predicate. This trajectory structure explains the similar frontiers despite the different predicate geometry.

\noindent\textbf{Contact monitoring.}
A contact-only monitor cannot detect a hazard defined at positive clearance. On the fixed benches, two thirds of the scored hazard events occur without contact with any object. A contact monitor using the same state information intervenes on 9.7\% of actions but leaves the hazard rate unchanged, losing 17 successful episodes while gaining only two. The task-success reduction is statistically significant at $p=7.3\times10^{-4}$, whereas the change in hazard rate is not. On the corresponding random-layout evaluation, the contact monitor yields a hazard rate of 0.208 compared with 0.198 without it. Contact therefore provides neither sufficient coverage of positive-clearance events nor an effective proxy for the proposed geometric predicate.

\noindent\textbf{Computational overhead.}
Each predicate evaluation requires one geometry-to-geometry distance query per terminal and forecast step. With two terminals and $H=3$, this gives six distance queries per control step, obtained from the simulator broadphase. This cost is negligible relative to a forward pass through the 7B policy. In a physical system, the dominant additional cost would depend on how terminal and payload clearances are estimated.

\section{Discussion}

\noindent\textbf{Fallback Strategy.}
Our results show that post-intervention behavior is a major determinant of the safety--utility trade-off. With the predicate and margin fixed, retreat preserves substantially more task success than holding, while both minimally invasive strategies leave considerably more residual hazard without a measurable gain in task success. This behavior is consistent with the geometry of the hazard. A correction that remains close to the policy command can keep the payload near the active constraint boundary, whereas retreat explicitly moves it away. For proximity-defined hazards, minimizing action deviation is therefore not necessarily aligned with minimizing residual risk.

\noindent\textbf{Margin-Dependent Predicate Comparison.}
Predicate structure should be evaluated jointly with its margin over a range of operating points. The conjunctive predicate, two-site keep-out, and composite predicate differ markedly at small fixed margins, but their safety--utility frontiers become similar when each is evaluated over its own range of margins. The second-policy experiment further shows that margins selected for one payload may require retuning for another. These results indicate that margin selection is an integral part of filter design.

\noindent\textbf{Benchmark Design.}
Hazard geometry does not determine whether a learned policy encounters the unsafe region. In our experiments, reachability depends on whether the nominal transport trajectory crosses the terminal pair and at what height. Contact-based monitoring is insufficient because many scored events occur at positive clearance. Evaluations of runtime filters for this hazard class should therefore report the task trajectory, terminal geometry, evaluation allowance, and filter margin. Without these factors, differences between filtering methods can be confounded by whether the policy encounters the hazard in the first place.

\section{Limitations}
This study has several limitations. Because experiments with energized battery hardware require dedicated safety infrastructure, we evaluate runtime filtering in LIBERO using a geometric clearance criterion instead of a physical battery-disassembly system or an electrical short-circuit model. The filter receives terminal and payload geometry directly from the simulator, while our robustness study considers persistent clearance errors without a complete perception pipeline. Our evaluation focuses on empirical runtime intervention, with the barrier projection solved locally and without formal safety certification. Future work can extend this study to physical battery systems, perception-based geometry estimation, certified filtering, and broader workcells and policies where predicate geometry may play a larger role.

\section{Conclusion}

We studied runtime safety filtering for a two-terminal manipulation hazard whose unsafe region depends on simultaneous proximity to two terminals. By separating predicate structure, margin, and fallback strategy, we find that the conjunctive predicate, conventional two-site keep-out, and their composite achieve similar safety--utility frontiers once each is evaluated over appropriate margin values. In contrast, the fallback strategy has a substantially larger effect: retreat consistently provides a stronger balance between task success and residual hazard than holding or the two minimally invasive alternatives. These trends persist under a second policy, additional workcell conditions, and biased clearance estimates. Overall, the results indicate that effective runtime filtering for proximity-defined hazards depends not only on how the unsafe set is represented, but also on how the intervention margin is selected and how the robot responds after an action is rejected.

\bibliographystyle{IEEEtran}
\bibliography{references}

\end{document}